\pdfoutput=1

\documentclass[11pt]{article}

\usepackage{booktabs}
\usepackage{longtable}
\usepackage{makecell}
\usepackage{amssymb}
\usepackage{graphicx}
\usepackage{bbding}
\usepackage{tabularx}
\usepackage{csquotes}
\usepackage{longtable}

\usepackage[preprint]{acl}

\usepackage{times}
\usepackage{latexsym}

\usepackage[T1]{fontenc}

\usepackage[utf8]{inputenc}

\usepackage{microtype}

\usepackage{inconsolata}

\usepackage{xspace}
\newcommand{\method}{\textsc{DeepEval}}
\newcommand{\cpm}[1]{\textcolor{gray}{$_{\pm #1}$}}
\title{Can Large Multimodal Models Uncover Deep Semantics Behind Images?}

\author{\textbf{{Yixin Yang}\textsuperscript{\rm $\dagger$}, {Zheng Li}\textsuperscript{\rm $\dagger$}, {Qingxiu Dong}\textsuperscript{\rm $\dagger$}, {Heming Xia}\textsuperscript{\rm $\diamond$}, {Zhifang Sui}\textsuperscript{\rm $\dagger$$\ddagger$}} \\
\textsuperscript{\rm $\dagger$} State Key Laboratory of Multimedia Information Processing, Peking University\\
  \textsuperscript{\rm $\diamond$} Department of Computing, The Hong Kong Polytechnic University \\ 
  \textsuperscript{\rm $\ddagger$} Jiangsu Collaborative Innovation Center for Language Ability, Jiangsu Normal University\\
  \tt \{yangyx,dqx\}@stu.pku.edu.cn, \{lizheng2001, szf\}@pku.edu.cn \\
  \tt he-ming.xia@connect.polyu.hk}

\begin{document}
\maketitle
\begin{abstract}
Understanding the deep semantics of images is essential in the era dominated by social media. 
However, current research works primarily on the superficial description of images, revealing a notable deficiency in the systematic investigation of the inherent deep semantics. 
In this work, we introduce \method{}, a comprehensive benchmark to assess Large Multimodal Models' (LMMs) capacities of visual deep semantics.
\method{} includes human-annotated dataset and three progressive subtasks: fine-grained description selection, in-depth title matching, and deep semantics understanding.  
Utilizing \method{}, we evaluate 9 open-source LMMs and GPT-4V(ision). 
Our evaluation demonstrates a substantial gap between the deep semantic comprehension capabilities of existing LMMs and humans. 
For example, GPT-4V is 30\% behind humans in understanding deep semantics, even though it achieves human-comparable performance in image description. 
Further analysis reveals that LMM performance on \method{} varies according to the specific facets of deep semantics explored, indicating the fundamental challenges remaining in developing LMMs.\footnote{The dataset and code for the experiments are available at: https://github.com/AnnaYang2020/DeepEval.}


\end{abstract}

\section{Introduction}

\begin{quote}\textit{The image is more than an idea. It is a vortex or cluster of fused ideas and is endowed with energy.}\par\raggedleft--- Ezra Pound (1915)\end{quote}  
Deep semantics of an image refer to the underlying meanings that extend beyond the superficial interpretation, probing into the essence of the image~\cite{barthes1968elements}. 
Although not every image inherently carries profound semantics, the concept of deep semantics is widespread across various fields~\cite{barthes1999rhetoric,deman2010comics,barthes2000photographic,somov2005semiotic,somov2006connotations}.
Understanding the deep semantics of images is a manifestation of high-level human intelligence, serving as an important means of exploration from perceptual intelligence to cognitive intelligence.
\begin{figure}[t]
\centering
\includegraphics[width=0.95\columnwidth]{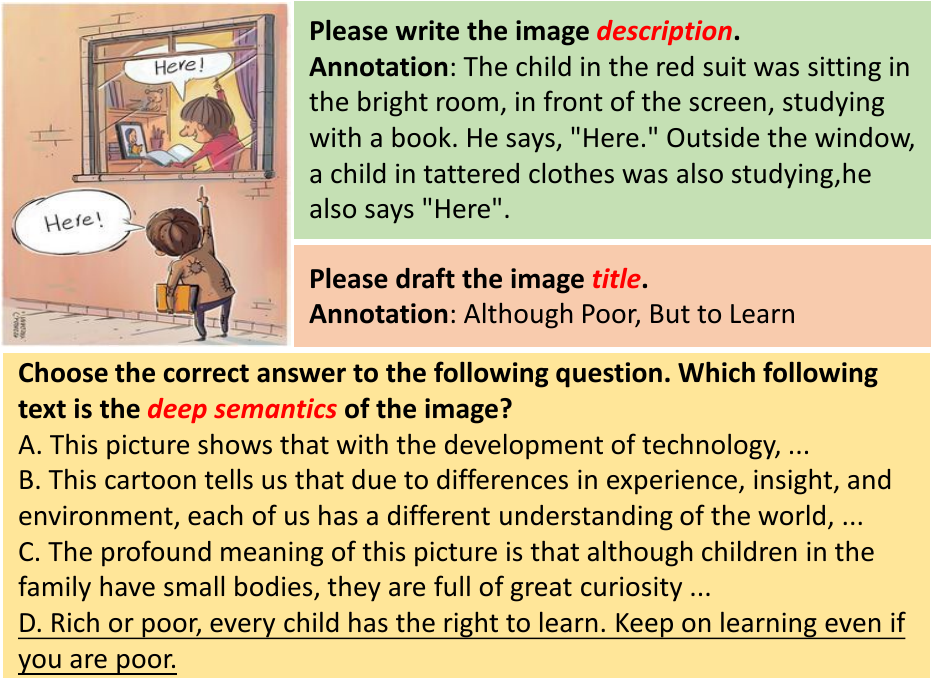}
\caption{An example from the \method{} dataset includes annotated description, annotated title, and the corresponding multiple-choice question for deep semantics from the Deep Semantics Understanding Task.}
\label{fig:fig_sim}
\end{figure}

However, previous efforts in visual understanding mainly focus on surface-level aspects of images, such as object attributes~\cite{wang2022co} 
and relationship reasoning~\cite{hudson2019gqa}.
Earlier attempts on deep semantic are limited in scope, focusing solely on sarcasm or humor,
~\cite{cai2019multi,chauhan2022emoji,boccignone2017amhuse,patro2021multimodal}, 
and lack in systematic investigation of the inherent deep semantic.

To address the mentioned limitations and fill the current research gap, we introduce \method, a benchmark for understanding the deep semantics of cartoons across various categories, accompanied by a meticulously annotated dataset. Additionally, we introduce three tasks: Fine-grained Description Selection, In-depth Title Matching, and Deep Semantics Understanding, to comprehensively evaluate models' capabilities in understanding deep semantics. Cartoons, often imbued with profound meanings by their creators, are an ideal subject for this study. The \method{} dataset comprises over 1,000 samples, each featuring a cartoon image and manually annotated components, including image description, title, and deep semantics. Moreover, we develop multiple-choice questions for quantitative assessment, tailored for each task.

We conduct evaluations on various open-source LMMs as well as the proprietary GPT-4V(ision). Our findings reveal a significant gap between the capabilities of AI and humans in understanding deep semantics. Models with a larger number of parameters generally demonstrate a better understanding of deep semantics. Moreover, we discover that incorporating a description significantly helps these models in grasping the underlying semantics of an image. Furthermore, We also explore the performance of different models across various categories of images. By undertaking \method{}, our goal is to promote research in model development, focusing on a deeper understanding of semantics in visual content.

\section{Related Work}
\paragraph{Large Multimodal Models}
Large language models (LLMs) have exhibited strong abilities in various natural language understanding and generation tasks~\cite{Hugo:2023llama, touvron2023llama2,ray2023chatgpt}. Drawing on LLMs' scaling law, a series of Large Multimodal Models (LMMs) using LLMs as the backbone has emerged. These models~\cite{tsimpoukelli2021multimodal, Li2023BLIP2BL,liu2023llava, liu2023improvedllava, zhu2023minigpt, wang2023cogvlm, ye2023mplugowl2} have aligned visual features with language models through additional layers or specialized modules. Additionally, several closed-source LMMs~\cite{alayrac2022flamingo, driess2023palm}, especially GPT-4V~\cite{yang2023dawn}, show remarkable ability in managing complex multimodal inputs. These models have set new benchmarks in performance~\cite{fu2023mme,li2023seed}, increasingly becoming predominant in visual-language research. However, relevant studies suggest that LMMs still face limitations in comprehending deeper semantics~\cite{liu2023mmbench, yang2023mm}. 

\begin{figure}[t]
\centering
\includegraphics[width=1.0\columnwidth]{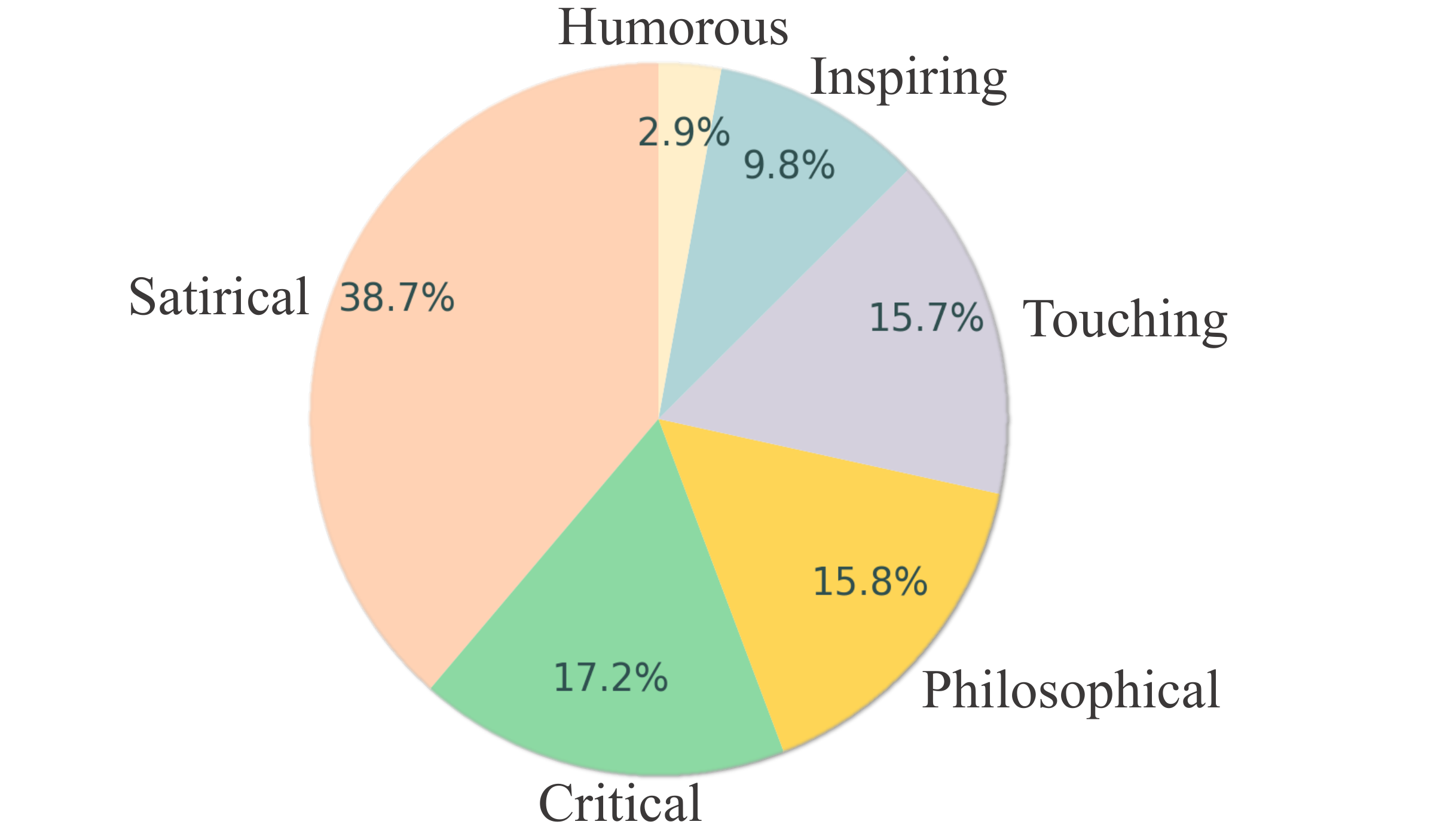}
\caption{The distribution of six categories of \method{} dataset.}
\label{fig:cate_dis}
\end{figure}

\paragraph{Visual Deep Semantics Understanding}
Understanding the deep semantics of visual contents represents a critical cognitive ability in humans. For AI, this ability showcases its depth of understanding images~\cite{9428534,kruk-etal-2023-impressions}. Present evaluations~\cite{Lin2014MicrosoftCC, antol2015vqa, goyal2017making, gurari2018vizwiz, hudson2019gqa, wang2022co, xia-etal-2023-imagenetvc} mainly concentrate on superficial aspect of understanding. Pioneering works in affective image classification~\cite{4711701, Machajdik2010AffectiveIC} have shown that LMMs are capable of attaining an understanding beyond mere surface content. Research in sarcasm~\cite{das2018sarcasm, cai-etal-2019-multi,lemmens-etal-2020-sarcasm,abu-farha-etal-2022-semeval} and humor detection~\cite{radev-etal-2016-humor} only employs classification tasks. Further work~\cite{desai2021nice} provides explanations for satirical content. The most relevant prior work~\cite{Hessel2022DoAL} selects humorous captions for images and provides explanations. However, it exclusively focuses on humor evaluation. In contrast, our work is pioneering in its comprehensive exploration of visual deep semantics across multiple categories, offering a more thorough assessment of the deep semantics within images. We provide a detailed comparison between our method and previous studies in Table~\ref{tab:Features}, and the categories covered by our method are illustrated in Figure~\ref{fig:cate_dis}.

\begin{table*}[t]
\centering
\small
\setlength{\tabcolsep}{0.6mm}{
\begin{tabular}{lcccccc}
\toprule
      \multicolumn{1}{l}{\textbf{Benchmark}} & \multicolumn{1}{c}{\textbf{Task}} & \multicolumn{2}{c}{\textbf{Semantics}}  & \multicolumn{1}{c}{\textbf{\# Category}} & \multicolumn{1}{c}{\textbf{Img Type}} \\ 
       \multicolumn{1}{c}{} & \multicolumn{1}{c}{} &  \multicolumn{1}{c}{\textbf{avg. length}} & \multicolumn{1}{c}{\textbf{size}} &  \multicolumn{1}{c}{}\\ 
\midrule
HCD~\cite{radev-etal-2016-humor} & Funniness Classification & -  & - & 1 & \begin{minipage}{0.3\columnwidth}
		\centering
		\raisebox{-.1\height}{\includegraphics[width=0.3\linewidth]{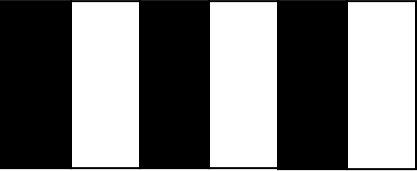}}
	\end{minipage} \\
FSD~\cite{das2018sarcasm} & Sarcasm Classification & - & -  & 1 & \begin{minipage}{0.3\columnwidth}
		\centering
		\raisebox{-.1\height}{\includegraphics[width=0.3\linewidth]{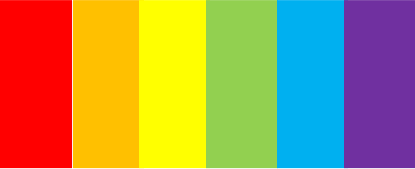}}
	\end{minipage}  \\
MTSD~\cite{cai-etal-2019-multi} & Sarcasm Classification & -  & - &  1 & \begin{minipage}{0.3\columnwidth}
		\centering
		\raisebox{-.1\height}{\includegraphics[width=0.3\linewidth]{Color.png}}
	\end{minipage} \\
RTSD~\cite{lemmens-etal-2020-sarcasm} & Sarcasm Classification & -  & - &  1 & \begin{minipage}{0.3\columnwidth}
		\centering
		\raisebox{-.1\height}{\includegraphics[width=0.3\linewidth]{Color.png}}
	\end{minipage} \\
iSarcasmEval~\cite{abu-farha-etal-2022-semeval} & Sarcasm Classification & -  & - &  1 & \begin{minipage}{0.3\columnwidth}
		\centering
		\raisebox{-.1\height}{\includegraphics[width=0.3\linewidth]{Color.png}}
	\end{minipage}\\
MORE~\cite{desai2021nice} & Sarcasm Explanation &  15 & 3510 &  1 & \begin{minipage}{0.3\columnwidth}
		\centering
		\raisebox{-.1\height}{\includegraphics[width=0.3\linewidth]{Color.png}}
	\end{minipage} \\
HUB\cite{Hessel2022DoAL}  & Matching+Ranking+Explanation &  60 & 651 & 1 & \begin{minipage}{0.3\columnwidth}
		\centering
		\raisebox{-.1\height}{\includegraphics[width=0.3\linewidth]{BW.png}}
	\end{minipage} \\
\toprule
\method{} (Ours)  &  Description+Title+\textbf{Deep Semantics} &  37 & 1001 &  6 & \begin{minipage}{0.3\columnwidth}
		\centering
		\raisebox{-.1\height}{\includegraphics[width=0.3\linewidth]{Color.png}}
	\end{minipage} \\
\bottomrule
\end{tabular}}
\caption{Features and statistical information of \method{} and prior related datasets. "Semantics" refers to the explanatory texts in More and HUB, as well as annotated deep semantics texts in our dataset. The term "ave. length" refers to the average length of sample texts, while "size" indicates the number of semantic text samples in the dataset. "Img Type" includes black and white images and color images. The "-" refers to no semantics text in classification task.}
\label{tab:Features}
\end{table*}

\section{Dataset and Task Overview}

The \method{} dataset includes 1,001 samples, each with an image and three manually annotated components: a description, a title, and deep semantics. The statistical information about the text is displayed in Table \ref{tab:statistics}. To enable quantitative evaluation, we additionally craft multiple-choice questions to test the understanding of descriptions, titles, and deep semantics. Each segment is represented by 1,001 questions, where each question presents an image, a question text, and four potential answers. Only one answer is correct, while the others serve as distractors. Figure~\ref{fig:fig_sim} illustrates examples of the manually annotated components and the multiple-choice questions.

To explore the capabilities of LMMs in comprehending the deep semantics of image, we construct a comprehensive evaluation consisting of three main subtasks:

\begin{itemize}
     \item \textit{Fine-grained Description Selection Task:} Evaluating the ability of models to accurately identify the surface-level details of images.
      \item \textit{In-depth Title Matching Task:} Assessing the capability of models to understand the overall signification of images.
      \item \textit{Deep Semantics Understanding Task:} Evaluating the ability of models to understand the detailed deep semantic meanings of images.
\end{itemize}

Together, these subtasks offer a robust and multifaceted evaluation of LLMs, enabling a deeper understanding of their strengths and limitations in image understanding.

\begin{table}[t]

\centering
\small
\setlength{\tabcolsep}{5mm}{
\begin{tabular}{cccc}
\toprule
      \multicolumn{2}{c}{Dataset Size} & \multicolumn{2}{c}{Description Length}\\ 
      \multicolumn{2}{c}{} & tot. & \multicolumn{1}{c}{avg.} \\ \midrule
\multicolumn{2}{c}{1001} & 49,595 & 49.55\\
\toprule
    \multicolumn{2}{c}{Deep Semantics Length} & \multicolumn{2}{c}{Title Length}\\
    \multicolumn{1}{c}{tot.} & \multicolumn{1}{c}{avg.} & tot. & avg.\\ \midrule
 37,002  & 36.97 & 5,709 & 5.70 \\
\bottomrule
\end{tabular}}
\caption{Statistics of \method{} dataset. The length of the text is calculated by counting the number of words contained in the text.}
\label{tab:statistics}
\end{table}\label{tab:data_stats}

\section{Dataset Construction}

\begin{figure*}[t]
\centering
\includegraphics[width=0.9\textwidth]{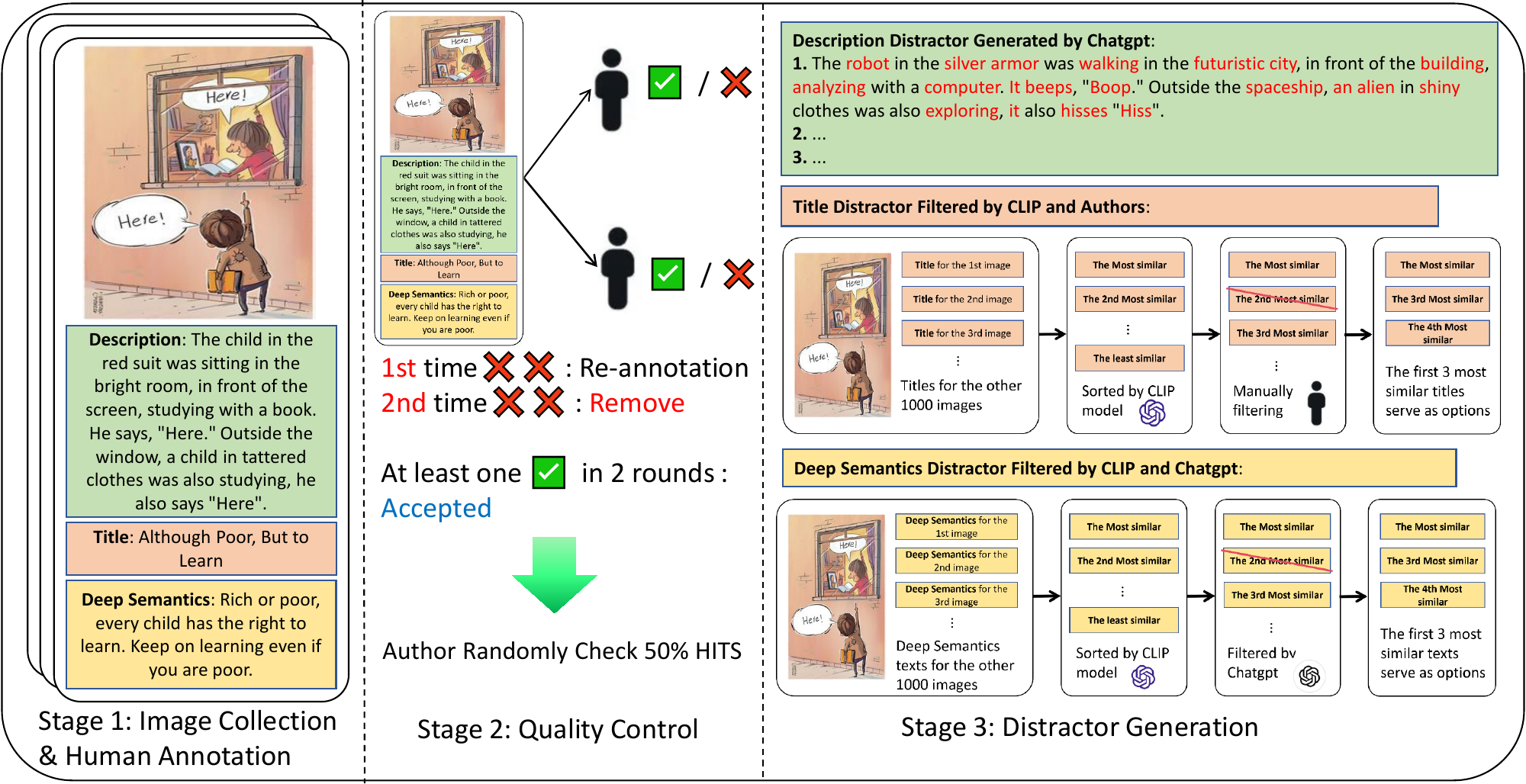}
\caption{Schematic diagram of \method{} dataset construction process including three stages: Image Collection \& Human Annotation, Quality Control and Distractor Generation.}
\label{fig:twocolumn}
\end{figure*}

We collect \method{} dataset in a multi-step crowd-sourcing pipeline, including 1) image collection, 2) data annotation, 3) option generation. With selected high-quality cartoon images, we ask crowd-source workers to write a description, a title and deep semantics of each image. 

\subsection{Image Collection}

The image data in the \method{} dataset are obtained by web scraping from websites. A total of 1,001 images are collected from Pinterest\footnote{https://www.pinterest.com/}, Cartoon Movement\footnote{https://cartoonmovement.com/}, and Google search. The gathered images span a diverse array of genres, encompassing satirical representations of current events, philosophical narratives, humorous and entertaining content, among others. After collection, a manual screening process is conducted to remove duplicates and unclear images.

\subsection{Data Annotation}

Deep semantics of images often requires extensive commonsense knowledge and advanced reasoning abilities. To obtain high-quality image descriptions, titles, and deep semantics, we primarily utilize manual annotation to collect gold-standard answers with rigorous quality controls. 

\subsubsection{Annotator Recruitment and Instruction}

We post a job description on online forums to invite over 50 applicants with at least a Bachelor's degree to participate in an online pre-annotating instruction and qualification test. Based on their preferences, we divide them into two groups: annotators and inspectors. After completing the pre-annotating instructions, we conduct a qualification test for quality control. In the end, we select 26 annotators and 18 inspectors.

\subsubsection{Cross-checking Annotation}

We divide the annotation process into 3 phases. In the first phase, annotators randomly select cartoon images from the dataset for annotation of image description, title and deep semantics. The image description and deep semantics should be over 80 characters, while the proposed title should be over 3 characters, or else they cannot be submitted. Subsequent to this phase, each image is transformed into a quadruple (image, description, title, deep semantics), marking the completion of the initial dataset construction.

In the second phase, inspectors will check the annotated images. When encountering low-quality annotations, they can drop them. Each image annotation is checked by two inspectors. If both inspectors drop the annotation, we drop the annotation and put the image back into the dataset for second annotation. 
If a comic image is rejected in two rounds of annotation, it indicates that the deep semantics conveyed by the image are unclear, so we will drop the image. During this stage, we also use Cohen's kappa to quantify the agreement between annotators, obtaining an average score of 0.701 across all tasks, which indicates substantial agreement~\cite{landis1977measurement}.

In the third phase, the authors further check all of the HITS from the second phase to ensure that the annotations meet our standards. Finally, we acquire 1,001 high-quality data entries, each represented as a quadruple (image, description, title, deep semantics).

\subsection{Options Generation}

After obtaining the image annotations, we use the annotated text as the correct option and construct three distractor options. Considering the high cost of constructing all distractor options using manual annotators, we utilize the power of CLIP model and ChatGPT model in this section.

For image descriptions, we employ ChatGPT model to generate sentences that retain their original sentence structure but alter the nouns, verbs, adverbs, or adjectives. This generates more intrusive options in fine-grained description selection task. Then, the authors manually check all the options to ensure that the multiple-choice questions maintain a certain level of difficulty while having a unique and correct answer. Detailed prompts and examples can be found in Appendix ~\ref{sec:des-options-eg}.

For deep semantics of the image, we use the CLIP model to calculate the similarity between the image and other deep semantics texts. We aim to select texts with higher similarity scores as distractors to create more challenging options. However, due to the presence of images with similar themes in the dataset, which may share similar semantics and potentially cause confusion, we utilize the ChatGPT model to eliminate distractor texts that are too similar to the correct option. Subsequently, we select the top three terms with the highest similarity as distractor terms.

For image titles, we similarly utilize the CLIP model to determine the similarity between the image and other titles. However, since there can be numerous titles with distinct meanings that might serve as the title for the same image, determining whether a title causes confusion becomes more challenging. Therefore, in this part, the authors manually filter out confusing distractor texts and select texts with high similarity scores as distractor options.

\subsection{Subtask Composition}

We divide the task of understanding the deep semantics of cartoon into three progressive subtasks: fine-grained description selection, in-depth title matching, and deep semantics understanding. Among them, the fine-grained description selection task requires multi-modal models to identify the surface-level details of the images. The in-depth title matching task requires models to comprehend the overall significance of the images and grasp their basic intentions. The deep semantics understanding task takes it a step further by demanding multi-modal models to acquire a comprehensive and detailed understanding of the thoughts, connotations, and information conveyed in the images. It can be observed that these three tasks gradually augment the comprehension of images, each task building upon the previous one to deepen the level of understanding. In these three tasks, each question consists of an image and a multiple-choice question with four options. The model is then required to select the option it believes best conveys the description, title, or deep semantics from the four options.

\subsection{Dataset Quality}

To ensure the quality of the dataset, the authors have checked all the data for descriptions, titles, deep semantic annotations, and the multiple-choice questions of the three tasks. This ensures that the content of descriptions, titles, and deep semantics annotations meet the standards and maintain high quality. For the multiple-choice questions, this confirms that they are challenging and contain standard answers. Furthermore, we employ annotators to evaluate the triplets of each image (description, title, deep semantics) and provide a score between 1 and 5. A score of 1 indicates complete inconsistency, a score of 5 indicates complete consistency, and each image is evaluated by three different annotators. Finally, our dataset obtained an average score of 4.74, indicating that our dataset is of high quality.

\subsection{License and Copyright}

In this dataset, we used original web links of comic images without infringing on their copyright. For images sourced from MathPile governed by licenses stricter than CC BY-NC-SA 4.0, MathPile adheres to the more restrictive licensing terms. Otherwise, it operates under the CC BY-NC-SA 4.0 license. This work is licensed under a CC BY-NC license. Our annotators participate in the annotation process voluntarily and receive fair compensation.

\section{Experiments}

\begin{table*}[t]
\centering
\small
\setlength{\tabcolsep}{3mm}{
\begin{tabular}{lcclll}
\toprule
\textbf{Model} &\textbf{Backbone} &\textbf{\# Params} & \textbf{Description} &\textbf{Title} & \textbf{DeepSemantics}\\\midrule
CogVLM~\cite{wang2023cogvlm} &Vicuna-v1.5 &17B & 72.83\cpm{6.81} & 45.05\cpm{5.89} & 32.20\cpm{1.00} \\
InstructBlip-13B~\cite{Dai2023InstructBLIPTG} & Vicuna-v1.5 &14B &59.44\cpm{6.12} &36.66\cpm{3.55} & 15.75\cpm{2.04} \\
LLaVA-1.5-13B~\cite{liu2023improvedllava} & Vicuna-v1.5 & 13B & 53.91\cpm{10.92} & 35.13\cpm{5.16} & 25.71\cpm{0.16} \\
Qwen-VL-Chat~\cite{Qwen-VL} & Qwen &10B &78.82\cpm{4.68} & 47.68\cpm{1.79} &28.30\cpm{0.40} \\
mPlug-Owl2~\cite{ye2023mplugowl2} &LLaMA2 &8B &75.26\cpm{3.66} & 47.75\cpm{0.85} &31.37\cpm{2.55} \\
MiniGPT-4~\cite{zhu2023minigpt} &LLaMA2 &8B &41.79\cpm{5.74} &33.00\cpm{4.30} &26.34\cpm{2.24} \\
InstructBlip-7B~\cite{Dai2023InstructBLIPTG} &Vicuna-v1.5 &8B    &49.88\cpm{6.18} &32.23\cpm{4.87} &15.72\cpm{1.26} \\
Fuyu~\cite{fuyu-8b} &- &8B  &29.90\cpm{0.16} &26.54\cpm{0.36} &17.44\cpm{1.66} \\
LLaVA-1.5-7B~\cite{liu2023improvedllava} &Vicuna-v1.5 &7B &48.62\cpm{13.61} &32.00\cpm{6.48} &24.94\cpm{2.05} \\\midrule
GPT-4V~\cite{yang2023dawn} &-  &- &\textbf{96.53}\cpm{1.06} &  \textbf{55.01}\cpm{0.96} &\textbf{63.14}\cpm{3.00}\\\midrule
Human &-  &- &  100.00   &  94.00  & 93.00 \\
\bottomrule 
\end{tabular}}
\caption{The benchmark includes the average accuracy (in percentages (\%)) and variance on three prompts for the \method{} method. Description, Title and DeepSemantics represent Fine-grained Description Selection Task, In-depth Title Matching Task, and Deep Semantics Undertanding Task respectively.}
\label{tab:experiments}
\end{table*}
\subsection{Baselines}

In consideration of the strong performance exhibited by LMMs in addressing image comprehension challenges, we evaluate the following LMMs: LLaVA-1.5~\cite{liu2023improvedllava}, MiniGPT-4~\cite{zhu2023minigpt}, mPLUG-Owl2~\cite{ye2023mplugowl2}, CogVLM~\cite{wang2023cogvlm}, Qwen-VL~\cite{Qwen-VL}, InstructBlip2~\cite{Dai2023InstructBLIPTG}, Fuyu~\cite{fuyu-8b}, GPT-4V~\cite{yang2023dawn}. A detailed introduction to these models can be found in the Appendix ~\ref{sec:lmms}.

\subsection{Experiment Details}

In evaluating performance for our tasks, accuracy serves as the primary metric. A model's answer is considered correct when it aligns with the established standard answer. Accuracy is quantified by the ratio of the number of correct responses $N_{r}$, to the total number of question $N$, expressed as $N_{r}/N$. Our task prompts start with specifying  description, title, or deep semantics, followed by multiple-choice options: A, B, C, and D. To minimize deviations in results caused by variations in the text descriptions within the prompt, we develop three distinct prompt formats, which are elaborately described in Appendix~\ref{sec:prom-deta}. The parameters for each model used in the experiment, including possible settings for temperature and top-k, are comprehensively detailed in Appendix~\ref{sec:mod-para-de}. Furthermore, to assess human capabilities in these tasks, we randomly select 100 questions from the dataset for each task and have human evaluators answer them. This allows us to benchmark the performance of human participants against our models, offering a thorough comparison of both human and machine proficiency in these specific tasks.

\subsection{Main Results}
\paragraph{Fine-grained Description Selection Task}

The results of various LMMs in fine-grained description selection task are shown in Table \ref{tab:experiments}. It can be observed that Qwen-VL-Chat, among the open-source models, exhibit the highest recognition capability for fine-grained surface description, with an accuracy of 78.82\%. On the other hand, Fuyu demonstrates the weakest recognition ability for fine-grained surface-level information, with an accuracy of only 29.90\%. The latest GPT-4V exhibits outstanding performance with an impressive accuracy of 96.53\%. Nevertheless, these models still do not match the capabilities of humans, whose accuracy remains at a perfect 100\%. 

\paragraph{In-depth Title Matching Task}

The performance of the models in the in-depth title matching task is also presented in Table \ref{tab:experiments}. Among the open-source models,  mPlug-Owl2 performs the best with an accuracy of 47.75\%, while Fuyu shows the weakest performance with an accuracy of only 26.54\%. The closed-source model GPT-4V outperforms them all, achieving an accuracy of 55.01\%. A notable observation across all models is that their performance in this task significantly trails behind their performance in the preceding fine-grained description selection task. This indicates that processing deep semantics is more challenging, despite the in-depth title matching task primarily addressing the broad essence rather than intricate details of deep semantics. Additionally, it's evident that these models substantially fall short of human-level performance, which is marked at an impressive 94\%.

\paragraph{Deep Semantics Understanding Task}

Among open-source models, CogVLM showcases the highest performance with an accuracy of 32.20\%, while InstructBlip-7B scores the lowest, achieving only 15.72\% accuracy, shown in Table \ref{tab:experiments}. Unsurprisingly, GPT-4V achieves better results with an accuracy of 63.14\%. However, GPT-4V exhibits the largest variance among the evaluated models models in deep semantics understanding, indicating instability despite its overall superior performance. Furthermore, when comparing GPT-4V's results across all tasks, there is notably higher variance in the deep semantics aspect, suggesting weaker performance compared to other tasks. Additionally, we note that the capabilities of these models are significantly weaker than human performance, which stands at 93\%.

It can be observed that the accuracy of all evaluated models in deep semantics understanding is significantly lower than their performance in image description, and nearly all of them achieve lower accuracy in deep semantics understanding compared to the in-depth title matching task. This underscores that comprehending deep semantics of images presents a significant challenge for these models, and focusing on the finer details of deep semantics adds further complexity, aligning with our expectations. Interestingly, only GPT-4V demonstrates higher accuracy in the deep semantics task compared to the in-depth title matching task. This could suggest that GPT-4V's stronger understanding of longer texts, coupled with the detailed information conveyed in deep semantics texts, aids the model in making more accurate judgments in deep semantics understanding.
\begin{figure*}[t]
\centering
\includegraphics[width=\textwidth]{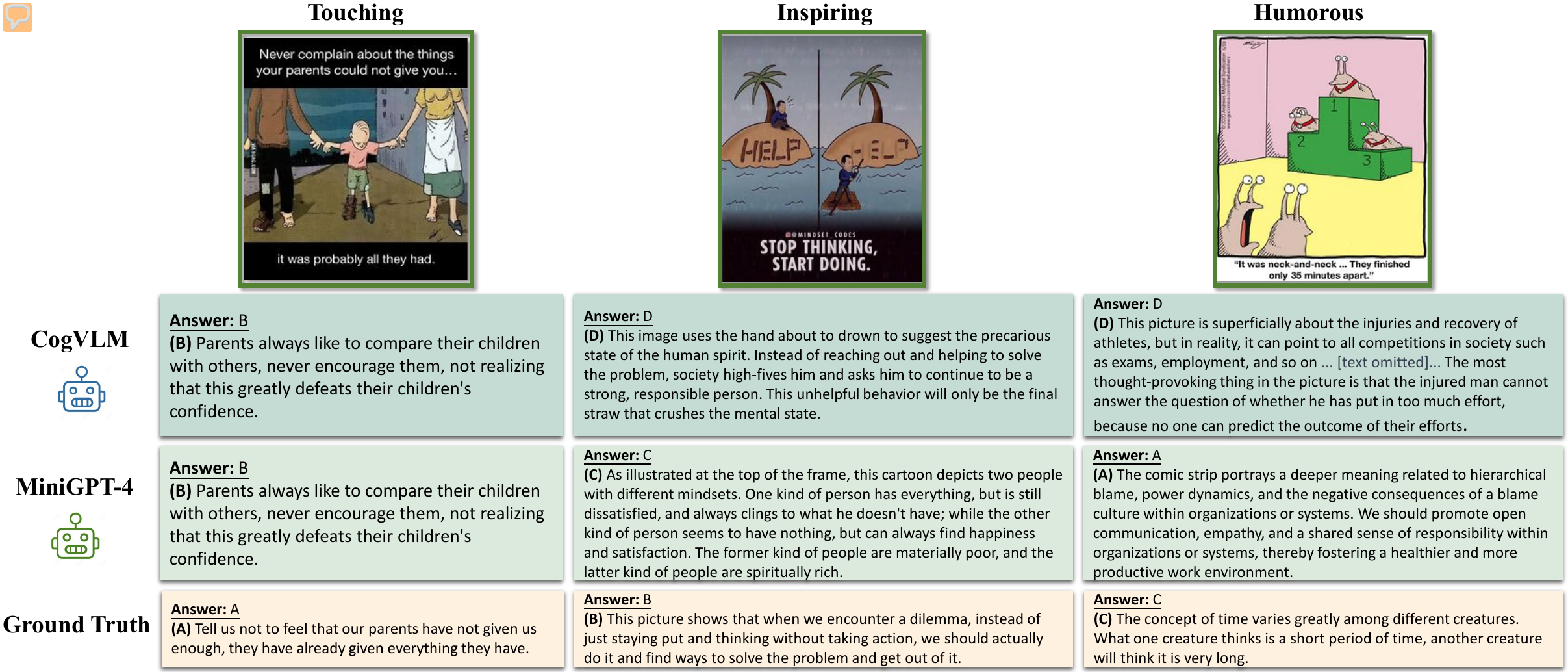}
\caption{Random samples of answers chosen by CogVLM and MiniGPT-4, along with the standard answers, covering three categories: \textit{Touching}, \textit{Inspiring}, and \textit{Humorous}, with one sample from each category.}
\label{fig:17}
\end{figure*}

\section{Analysis}

\subsection{How do models perform across various categories in image understanding?}
\begin{figure*}[t]
\centering
\includegraphics[width=0.85\textwidth]{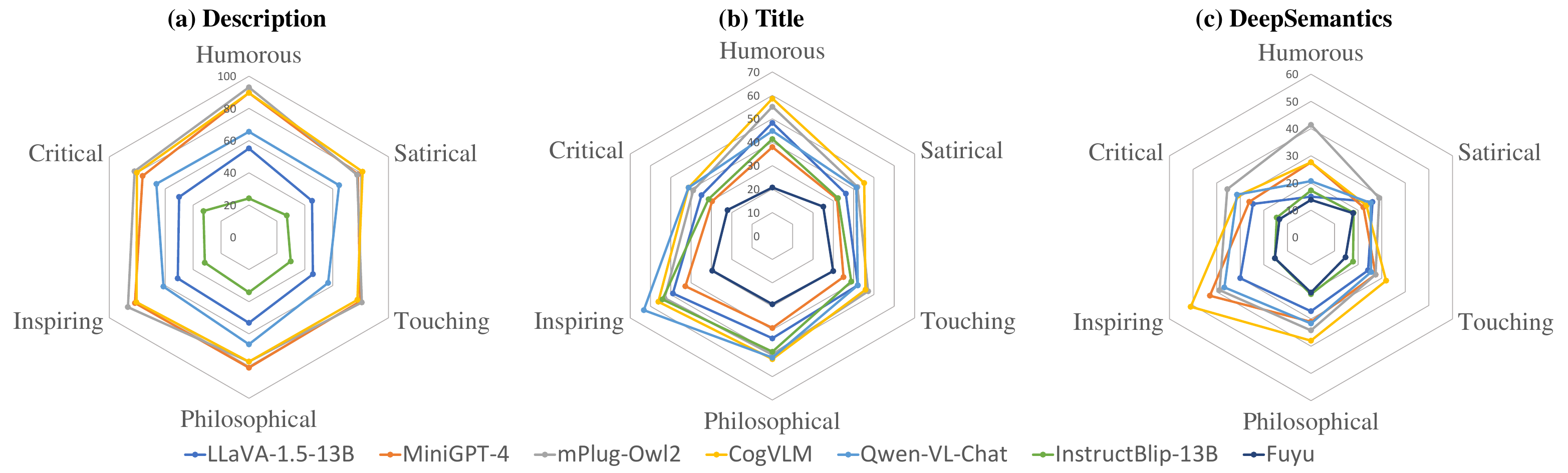}
\caption{The radar charts represent the performance of several typical models in understanding images across different categories in our three tasks.}
\label{fig:cate}
\end{figure*}

By analyzing the model's understanding capabilities in different categories, we can pinpoint strength or weakness of models in specific categories. The performance of different models across categories is illustrated in Figure~\ref{fig:cate}, with three radar charts showcasing the model's ability in interpreting image descriptions, titles, and deep semantics across different categories. The deep semantics graph reveals that different models exhibit their strengths in different categories. For instance, the mPlug-Owl2 and CogVLM stand out in the \textit{Humorous} and \textit{Inspiring} categories, respectively. Furthermore, despite extensive prior research, \textit{Satirical} category continues to challenge all models, with accuracy rates remaining below 30\%. This underscores the \textit{Satirical} category as a critical area for further research in understanding deep semantics within images. 

The description selection task's radar charts, resembling regular hexagons, indicate a more uniform comprehension of image descriptions across categories by the models. When evaluating titles, models show remarkable competency in both \textit{Humorous} and \textit{Inspiring} categories compared to others. However, regarding deep semantics, \textit{Inspiring} consistently emerges as the top-performing category for four models, whereas a majority struggle with \textit{Humorous}. This discrepancy may stem from the fact that \textit{Inspiring} content can often be summarized in few sentences. In contrast, \textit{Humorous} content typically involves more intricate interpretations that are heavily reliant on cultural context, timing, and the subtleties of language and expression. To provide a more intuitive display, Figure \ref{fig:17} showcase samples from typical categories in the deep semantics understanding task for CogVLM, MiniGPT-4, and the standard answers, while additional samples for other categories are available in Figure \ref{fig:16} in the Appendix \ref{sec:cate-sam}.

\subsection{Can image descriptions aid models' understanding of deep semantics?}
It is commonly believed that models need to first identify the content of image descriptions before further comprehending the deep semantics. Therefore, we are curious to explore whether inspiring the model by incorporating its surface image descriptions during the evaluation process would aid in the model's understanding of deep semantics. This process is divided into two steps: 1) having the model to generate detailed descriptions of the images; 2) incorporating the detailed descriptions into the prompt of the deep semantics understanding task. Additionally, to more effectively demonstrate the impact of integrating image descriptions on the understanding of deep semantics, we also directly include annotated image description texts in the prompt. In this case, the first step is omitted, and the detailed descriptions included in the second step are the annotated descriptions.

The results in Table \ref{tab:des} show that seven out of the nine evaluated models improve their understanding of deep semantics with model-generated image descriptions. These models had an average increase of 1.8 percentage points. Additionally, all nine models demonstrated better deep semantics understanding with annotated image descriptions, with an average increase of approximately 4.1 percentage points. Thus, incorporating the model's descriptions of the surface content appears to inspire and enhance its deep semantics understanding capabilities.

\begin{table}[t]

\centering
\small
\setlength{\tabcolsep}{1.5mm}{
\begin{tabular}{lccc}
\toprule
\textbf{Model}            & 
\textbf{\makecell{DS}}& \textbf{\makecell{DS \\(GeneDesc)}}& \textbf{\makecell{DS \\ (AnnoDesc)}}\\ 
 \midrule
 CogVLM          & 31.17 &  32.57    & 37.96    \\
 InstructBlip-13B      & 17.77  & 19.78 & 23.48    \\
 LLaVA-1.5-13B       & 25.88 & 26.87  & 30.07     \\
 Qwen-VL-Chat     & 28.37  &  28.17  & 34.57     \\
 mPlug-Owl2         & 31.97 &  35.46    & 41.16     \\
 MiniGPT-4       & 27.27 &  27.77    & 34.07 \\
 InstructBlip-7B      & 14.29  & 19.38   & 19.38 \\
 Fuyu            & 16.98  & 16.78  &  23.98  \\
 LLaVA-1.5-7B        & 27.27 & 30.83  & 30.07 \\
\bottomrule
\end{tabular}}
\caption{The model's capability to comprehend the deep semantics of images while incorporating various image descriptions. "DS" stands for "Deep Semantics", "GeneDesc" represents integration of model-generated image descriptions. "AnnoDesc" signifies integration of annotated image descriptions.} 
\label{tab:des}
\end{table}

\subsection{How does model parameter size affect deep semantics understanding?}
\begin{figure}[t]
\centering
\includegraphics[width=0.7\columnwidth]{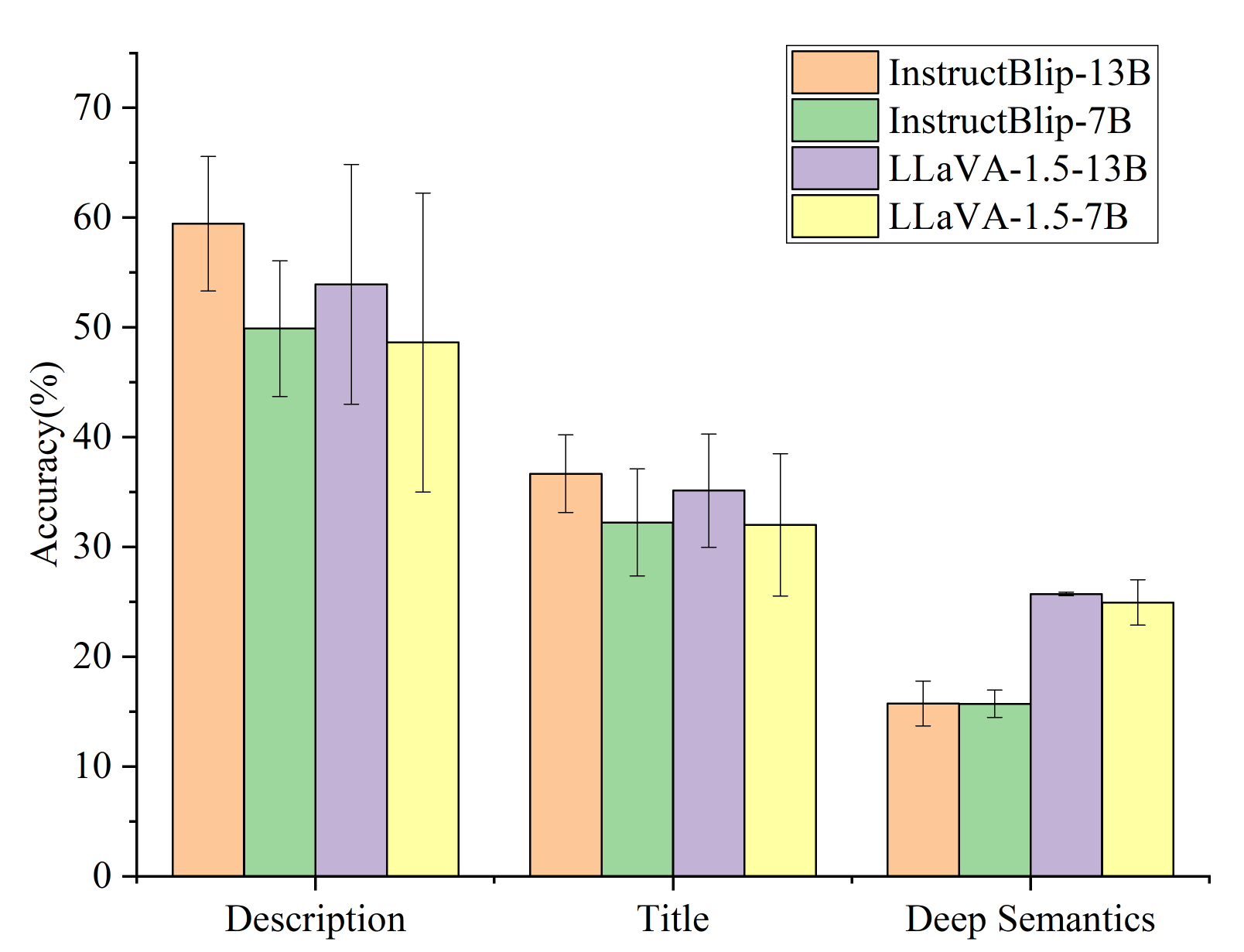}
\caption{Comparison of the average accuracy and variance results between InstructBlip-13B vs InstructBlip-7B and LLaVA-1.5-13B vs LLaVA-1.5-7B.}
\label{fig:par}
\end{figure}

Due to the scaling law, the number of parameters generally has a positive impact on the model's performance. In this context, we also discuss the relationship between model parameters size and deep semantics understanding. We examine two pairs of models, InstructBlip-13B vs. InstructBlip-7B and LLaVA-1.5-13B vs. LLaVA-1.5-7B, where each pair has consistent architecture and training processes, differing only in parameter size. Figure \ref{fig:par} provide a visual representation of the means and variances of accuracy across three tasks for these four models. It is observable that the 13B models have higher accuracy across all three tasks compared to the 7B models, indicating superior performance of the 13B models. Furthermore, the overall variances of the 7B models is higher than that of the 13B models. This indicates that, generally speaking, the 13B models are also more stable than the 7B models. Therefore, an increase in the number of parameters has a positive impact on the models' deep semantics understanding capabilities.

\section{Conclusion}
We propose \method{}, a benchmark for visual deep semantics of LMMs. \method{} consists of well-annotated dataset and three subtasks: fine-grained description selection, in-deep title matching, and deep semantic understanding. Evaluations are conducted on the leading LMMs, revealing a significant gap between AI and human capabilities in understanding deep semantics. Further analysis indicates that integrating image descriptions during the inference process  enhances LMMs’ ability to perceive deep semantics. The model's ability to understand deep semantics also improves as the number of parameters increases. Furthermore, our dataset is divided into multiple categories, and we conduct a more detailed analysis within these categories. Existing models still have a long way to go in terms of visual deep semantics understanding compared to humans. We hope that the proposed dataset and tasks can pave the way for AI to achieve a deeper understanding of the profound semantics conveyed by images.

\section*{Limitations}
The deep semantics of cartoon images are varied, and due to our limited collection of images, it's not feasible to encompass all potential deep semantic content. In this work, we have only exemplified some common categories, but the categories of images in the real world far exceed these six. On this note, adding more images and annotations would help improve this issue. 

Furthermore, our current images only include cartoons. This is because, compared to real-world pictures, cartoons generally contain rich and clear deep meanings, which are beneficial for investigating deep semantics. Despite this, our dataset images encompass a wide array of image types and a wide range of themes, reflecting the multifaceted nature of real-world scenarios. Detailed statistics for image types and themes can be found in Appendices \ref{sec:themes} and \ref{sec:images-type}. Our future work will expand to incorporate more types of images, such as photographs, advertising images, and artworks. 

Lastly, in the annotation process, we aim to reach a consensus among annotators on the deep semantics of images and only retain images with agreed-upon deep semantics. Therefore, images with deep semantics but significant controversy will not be included.

\section*{Acknowledgements}
This paper is supported by the National Key Research and Development Program of China (No.2020AAA0106700). The contact author is Zhifang Sui.

\bibliography{custom}

\appendix
\appendix

\section{Examples of Generating Distractor Generation For Description}
\label{sec:des-options-eg}
Considering that the generation of interference terms in the description only requires replacing nouns, adjectives, verbs, etc. in the sentence, we use Chatgpt to complete this task. The following is the prompt we use:\textit{Give me three different paragraphs that take only some of the verbs, nouns, adjectives, and adverbs in a given paragraph and modify words with irrelevant meanings. \\Input: [Example Input 1]\\Output: [Example Output 1]\\Input: [Example Input 2]\\Output: [Example Output 2]\\Input: [Example Input 3]\\Output: [Example Output 3]\\Input: [Input] \\Output:}

To ensure that ChatGPT understands our requirements correctly, we use a 3-shot prompt. These three examples were manually written by the author. The following is a modification example. It should be noted that the output of each example in the prompt has 3 modified paragraphs of text. For convenience, only one modified paragraph of text is shown here

\textbf{Source Text}: \textit{In the picture,there are three \textcolor{blue}{queues},the first one named Critic has many \textcolor{blue}{people},\textcolor{blue}{stand} in an endless \textcolor{blue}{line};the second one named Talker also has many \textcolor{blue}{people},but not that much as Critic;the third \textcolor{blue}{queue} named Doer,with no \textcolor{blue}{one in line}.}

\textbf{Revised Text}: \textit{In the picture, there are three \textcolor{red}{cats}, the first one named Critic has many \textcolor{red}{toys}, \textcolor{red}{play} in an endless \textcolor{red}{loop}; the second one named Talker also has many \textcolor{red}{toys}, but not that much as Critic; the third \textcolor{red}{cat} named Doer, with no \textcolor{red}{toys to play with}.}

\section{Prompt Details}
\label{sec:prom-deta}

To eliminate the influence of prompt expression on model performance, we used the following three types of prompts for testing:
\begin{itemize}
    \item \textit{Choose the correct answer to the following question. Which following text is the [description/best title/deep meaning] of the image? \\Options: (A) [...] (B) [...] (C) [...] (D) [...] \\Answer:}
    \item \textit{Select the appropriate [description/title/deep meaning] for the image from the options given. Which of these is the most suitable [description/title/deep meaning] for the image? \\Choices: A) [...] B) [...] C) [...] D) [...]\\Correct Answer:}
    \item \textit{Identify the most suitable [description/title/deep meaning] for the image from the given options. Which of the following should be chosen as the [description/title/deep meaning]? \\Choices are: A. [...], B. [...], C. [...], and D. [...]. \\The correct answer is:}
\end{itemize}

\section{Model Hyper-parameter Details}
\label{sec:mod-para-de}

We use the default hyper-parameter values of the models. In the LLaVa-1.5-7B and LLaVa-1.5-13B, the temperature is set to 0.2. For MiniGPT-4, the temperature is set to 1.0, and num\_beams is also set to 1.0. The temperature for mPlug-Owl-2 is set to 0.7. For CogVLM, the temperature is set to 0.4, top\_p is set to 0.8, and top\_k is set to 1.0.

\section{Categories Definition}
\label{sec:cate-def}

Table \ref{tab:intent} give the names and detailed definitions of the categories in \method.

\begin{table*}[!ht]
\caption{The names and specific definitions of the categories in \method.}
\label{tab:intent}
\centering
\small
\setlength{\tabcolsep}{3mm}{
\begin{tabular}{lp{10cm}}
\toprule
\textbf{Category}           & \textbf{Definition}\\
 \midrule
 Humorous & The image elicits amusement, laughter, or a sense of light-heartedness. It may contain elements that are funny, witty, or clever. \\
 \midrule
Critical & The image offers a critical perspective or analysis of a specific subject, aiming to examine and evaluate its merits, shortcomings, or implications. \\
\midrule
Touching & The image evokes strong emotions such as joy, sadness, empathy, or nostalgia. It may depict a heartwarming scene, a tender moment, or a poignant event. \\
\midrule
Philosophical & The image stimulates intellectual or philosophical contemplation. It raises questions, challenges assumptions, or encourages viewers to reflect on deeper meanings or concepts. \\
\midrule
Inspiring & The image motivates or uplifts viewers, conveying a positive message, encouraging resilience, or instilling hope. It may depict acts of kindness, achievement, or triumph over adversity. \\
\midrule
Satirical & The image conveys a message or commentary on a particular subject, often by using irony, sarcasm, or wit to highlight flaws or satirize societal norms, institutions, or individuals. \\
\bottomrule
\end{tabular}}
\end{table*}

\section{Categories Samples}
\label{sec:cate-sam}
\begin{figure*}
\centering
\includegraphics[width=\textwidth]{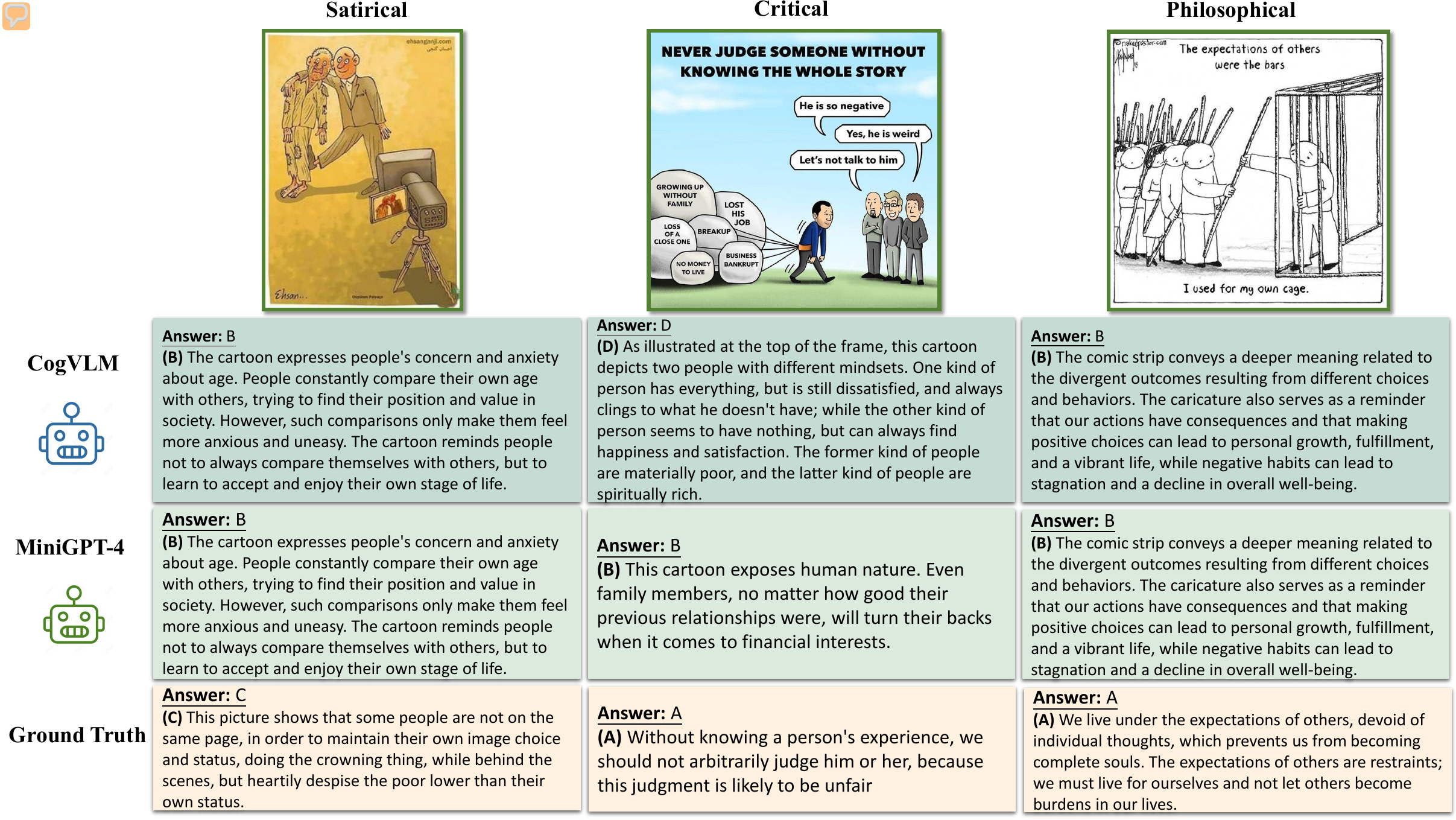}
\caption{Random samples of answers chosen by CogVLM and MiniGPT-4, along with the standard answers, covering three categories: \textit{Satirical}, \textit{Critical}, and \textit{Philosophical}, with one sample from each category.}
\label{fig:16}
\end{figure*}

Figure \ref{fig:16} give the samples of answers chosen by CogVLM and MiniGPT-4, in three \textit{Satirical}, \textit{Critical}, and \textit{Philosophical} category.

\section{Large Multimodal Models}
\label{sec:lmms}

\begin{itemize}

\item \textbf{LLaVA-1.5}~\cite{liu2023improvedllava} is an end-to-end LMM extended from Vicuna\cite{vicuna2023}, augmented with vision encoder.

\item \textbf{MiniGPT-4}~\cite{zhu2023minigpt} is an extension of Vicuna, incorporating ViT \cite{vit2021image} and Q-former \cite{li2023blip} as the vision encoder, while also featuring a single linear projection layer sandwiched between them.

\item \textbf{mPLUG-Owl2}~\cite{ye2023mplugowl2} is an extension of LLaMA-2-7B\cite{touvron2023llama2}, using ViT-L/14\cite{radford2021learning} as the vision encoder, and introducing a visual abstractor between them.

\item \textbf{CogVLM}~\cite{wang2023cogvlm} is also an extension of Vicuna, incorporating ViT\cite{vit2021image} as the vision encoder,  a two-layer MLP\cite{shazeer2020glu} as adapter, and introducing Visual expert module.

\item \textbf{Qwen-VL}~\cite{Qwen-VL} is an extension of Qwen-7B\cite{qwen}, incorporating ViT\cite{vit2021image} as the vision encoder, and introducing a vision-language adapter that compresses the image features.

\item \textbf{InstructBlip2}~\cite{Dai2023InstructBLIPTG} employs ViT-g/14 \cite{fang2022eva} as image encoder, and four different LLMs as language decoders. In our following tests, we utilize vicuna-13B and vicuna-7B \cite{vicuna2023} versions.

\item \textbf{Fuyu}~\cite{fuyu-8b} employs a decoder-only architecture, devoid of a dedicated image encoder for image processing. This design choice enables the model to support arbitrary image resolutions.

\item \textbf{GPT-4V}~\cite{yang2023dawn} is OpenAI's cutting-edge language model redefining natural language processing with advanced contextual understanding and versatile linguistic abilities.
\end{itemize}

\section{Wide Range of Themes} 
\label{sec:themes}
\begin{figure}[t]
\centering
\includegraphics[width=1.0\columnwidth]{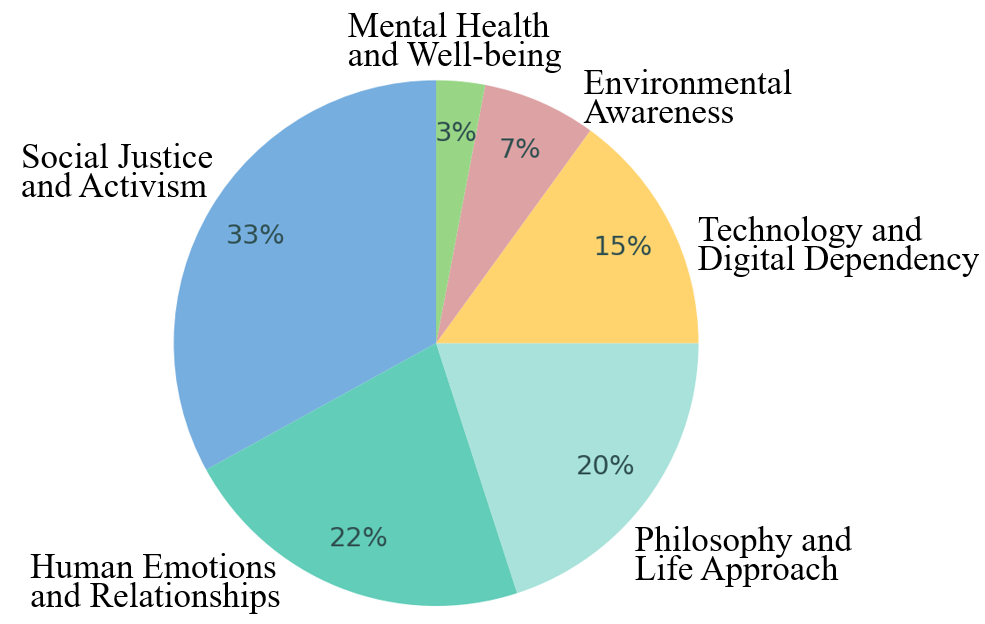}
\caption{The distribution of six themes of \method{} dataset.}
\label{fig:theme_dis}
\end{figure}

\method{} dataset boasts a wide range of themes, including Social Justice and Activism, Human Emotions and Relationships, Philosophy and Life Approach, Technology and Digital Dependency, Environmental Awareness, and Mental Health and Well-being. These diverse themes mirror the multifaceted nature of real-world scenarios. The corresponding statistics are detailed in Table \ref{fig:theme_dis}.

\section{Wide Array of Image Types} 
\label{sec:images-type}
\begin{figure}[t]
\centering
\includegraphics[width=1.0\columnwidth]{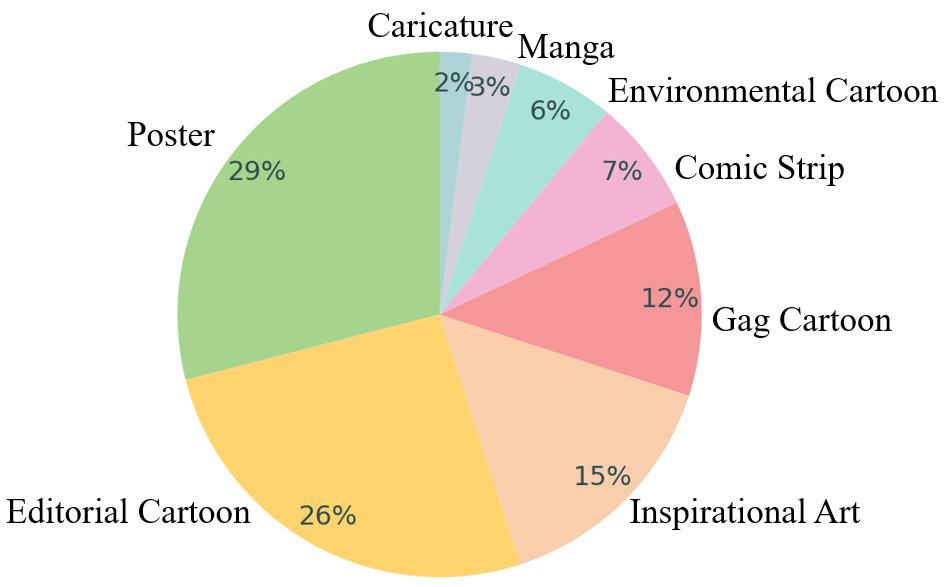}
\caption{The distribution of image types of \method{} dataset.}
\label{fig:type_dis}
\end{figure}

The term "cartoon" within \method{} dataset encompasses a wide array of image types, including Poster, Inspirational Art, Comic Strips, Manga, Caricature, Editorial Cartoon, and Environmental Cartoon. We have detailed statistics for these image types in Table \ref{fig:type_dis}.

\section{Influence of OCR Ability on Models’ Deep Semantics Ability} 
\label{sec:ocr-ability}

To study the potential impact of textual information within images on the models' understanding of deep semantics, we randomly selected 200 images and manually divided them into two groups: images without textual information (74 images) and images with textual information (126 images). We then calculated the mean accuracy and variance of each group across three distinct prompts to evaluate their deep semantic understanding capabilities. The results are shown in Table~\ref{tab:ocr}.

Our findings reveal that models demonstrate a higher ability to understand the deep semantics of images containing textual information compared to those without. This indicates a positive influence of OCR capabilities on the deep semantic understanding of images.

\begin{table*}[t]
\centering
\small
\setlength{\tabcolsep}{5mm}{
\begin{tabular}{lll}
\toprule
\textbf{Model} & \textbf{Img without Text} &\textbf{Img with Text}\\\midrule
CogVLM &30.63 \cpm{7.45} & 32.54 \cpm{2.86}  \\
InstructBlip-13B & 14.86 \cpm{2.34} & 16.93 \cpm{1.65}
\\
LLaVA-1.5-13B & 25.15\cpm{3.06} & 28.04\cpm{1.65} \\
Qwen-VL-Chat & 18.47\cpm{0.78} & 34.39\cpm{3.01} \\
mPlug-Owl2 & 21.62\cpm{0.00} & 33.07\cpm{3.20} \\
MiniGPT-4 &22.52 \cpm{2.06} & 25.66\cpm{6.42}\\
InstructBlip-7B & 14.41\cpm{3.40} & 15.87\cpm{2.10}\\
Fuyu &21.62\cpm {4.06} & 21.69\cpm{7.20}\\
LLaVA-1.5-7B &27.48\cpm{2.06} & 30.16\cpm{1.37}  \\
GPT-4V & 42.53\cpm{1.83} & 71.00\cpm{4.02} \\
\bottomrule 
\end{tabular}}
\caption{The model's capability to understand the deep semantics of images with and without textual information. The results includes the average accuracy (in percentages (\%)) and variance on three prompts for the \method{} method.}
\label{tab:ocr}
\end{table*}

\section{Generative Capabilities of Models} %
\label{sec:gene-result}
In addition to using multiple-choice questions to assess the model's capabilities, we further incorporate an additional assessment aimed at evaluating the generative performance on models with superior generative capabilities. To quantify the results, we use GPT-4 to determine the consistency between generated sentences and labeled sentences. Samples judged as consistent are considered correct, while those judged as inconsistent are considered incorrect. We then calculate the final accuracy. 

The results, detailed in Table~\ref{tab:gen}, reveal a close alignment between the outcomes of our main experiments and the generative capability assessments, thereby reinforcing our conclusions about the models' deep semantic understanding. Specifically, the Pearson coefficients between the results of main experiments and generative capability assessments are 0.95, 0.92, and 0.96 for the Description, Title, and DeepSemantics tasks, respectively.

\begin{table}[t]
\centering
\small
\setlength{\tabcolsep}{1.5mm}{
\begin{tabular}{llll}
\toprule
\textbf{Model} & \textbf{Description} &\textbf{Title} &\textbf{DeepSemantics}\\\midrule
CogVLM &80.32 & 46.95  & 31.07\\
LLaVA-1.5-13B & 68.93 & 39.66 & 26.77\\
Qwen-VL-Chat & 78.92 & 44.36 & 28.57\\
mPlug-Owl2 & 84.02 & 44.86 & 32.27\\
MiniGPT-4 &50.95 & 37.36 & 27.17\\
\bottomrule 
\end{tabular}}
\caption{The generative capabilities of models in three tasks. Description, Title and DeepSemantics represent Fine-grained Description Selection Task, In-depth Title Matching Task, and Deep Semantics Undertanding Task respectively.}
\label{tab:gen}
\end{table}

\end{document}